\documentclass[11pt,twocolumn,english,twocolumn]{article}
\usepackage[T1]{fontenc}
\usepackage[latin9]{inputenc}
\usepackage[a4paper]{geometry}
\usepackage{babel}
\usepackage{array}
\usepackage{graphicx}
\usepackage[unicode=true]
 {hyperref}

\makeatletter

\providecommand{\tabularnewline}{\\}
\newcommand{\lyxaddress}[1]{
\par {\raggedright #1
\vspace{1.4em}
\noindent\par}
}

\usepackage[font=small,labelfont=bf]{caption}
\makeatletter
\g@addto@macro\@floatboxreset\centering
\makeatother

\makeatother

\begin{document}
\twocolumn[   \begin{@twocolumnfalse}

\title{Molecular Generation with Recurrent Neural Networks (RNNs)}

\author{Esben Jannik Bjerrum\textsuperscript{1,{*}}Richard Threlfall\textsuperscript{2}}

\maketitle

\lyxaddress{1) Wildcard Pharmaceutical Consulting, Frødings Allé 41, 2860 Søborg,
Denmark. 2) Wiley-VCH, Boschstrasse 12, 69469 Weinheim, Germany.
{*}) \href{mailto:esben@wildcardconsulting.dk}{esben@wildcardconsulting.dk}}
\begin{abstract}
The potential number of drug like small molecules is estimated to
be between 10\textsuperscript{23} and 10\textsuperscript{60} while
current databases of known compounds are orders of magnitude smaller
with approximately 10\textsuperscript{8} compounds. This discrepancy
has led to an interest in generating virtual libraries by using hand
crafted chemical rules and fragment based methods to cover a larger
area of chemical space and generate chemical libraries for use in
\textit{in silico} drug discovery endeavors. Here it is explored to
what extent a recurrent neural network with long short term memory
cells can figure out sensible chemical rules and generate synthesizable
molecules through training on existing compounds encoded as SMILES.
The networks can, to a large extent generate novel, but chemically
sensible molecules. The properties of the molecules are tuned by training
on two different datasets consisting of fragment like molecules and
drug like molecules. The produced molecules and the training databases
have very similar distributions of molar weight, predicted logP, number
of hydrogen bond acceptors and donors, number of rotatable bonds and
topological polar surface area when compared to their respective training
sets. The compounds are for the most cases synthesizable as assessed
with synthetic accessibility (SA) score and Wiley ChemPlanner.
\end{abstract}
\bigskip{}

\hrule

\bigskip{}

 \end{@twocolumnfalse} ]

\section*{Introduction}

The number of potential drug like molecules is huge due to the large
flexibility and combinatorial potential of organic carbon, nitrogen
and oxygen chemistry. The number has been estimated to be between
10\textsuperscript{23} and 10\textsuperscript{60}\cite{Polishchuk2013,Ertl2003,Bohacek1996}
and dwarfs the current largest databases of chemical compounds, such
as ChemBL\cite{Bento2014}: \textasciitilde{}2{*}10\textsuperscript{6},
PubChem\cite{Kim2016} \textasciitilde{}90{*}10\textsuperscript{6}and
ChemSpider\cite{Pence2010} \textasciitilde{}60{*}10\textsuperscript{6}.
It has therefore long been of interest to generate virtual chemical
libraries for in silico drug discovery purposes\cite{Leach2000}.
A strategy for generation of a virtual library can be enumeration
of products from reaction of libraries of fragments, which ensures
the synthetic feasibility of the product and the availability of the
reaction fragments can be taken into account\cite{Hartenfeller2011}.
Examples of software solutions to enumerate such collections are eSynth\cite{Naderi2016}
and iLibDiverse built upon CombiGen\cite{Wolber2001}, and there are
also databases that have already been developed, such as the GDB sets\cite{Fink2007,Ruddigkeit2012,Blum2009}.
Many pharmaceutical companies have also developed solutions to enumerate
compounds that are within their synthetic reach and/or covered by
their current patents, thus defining a chemical space of compounds
with known synthetic routes and intellectual property coverage\cite{Nicolaou2016,Hu2012}.

Deep neural networks have gained a lot of interest for their ability
to do feature extraction and learn rules from presented training data.
The availability of graphics processing units (GPUs) and CUDA enabled
back-ends makes prolonged training available at a modest cost. New
regularization schemes such as dropout\cite{Srivastava2014} and noise
layers\cite{Luo2014} has enabled larger and deeper network architectures
without extensive over fitting. Task oriented architectures such as
convolutional neural networks\cite{Simard2003} (CNN) for spatial
related data such as images, have led to improvements in image analysis
whereas recurrent neural networks\cite{Chung2014} (RNNs)
have been successful for sequence based input. In RNNs, the state
of the network is propagated forward for each step of the input sequence,
enabling the network to alter its internal state for each input step
and thus alter the output even though the weights are the same in
the network. The performance of RNNs improved greatly with use of
micro architectures such as long short term memory (LSTM)\cite{Hochreiter1997}
cells and gated recurrent units (GRU)\cite{Chung2014}. LSTM cells
are a pre configured micro architecture of a neural network, which
has controlled gates for input, forget and output. This enables the
unit to keep its internal state for longer stretches of sequential
input in the RNN, leading to an improvement of the RNN performance.
The cells can be combined and stacked into architectures that have
interesting properties with regard to analysis of textual or sequence-based
inputs.

An interesting property of RNN\textquoteright s is their ability to
be played forward and generate new sequences\cite{Sutskever2011}.
This is done in an iterative manner where the predicted character
or next step in the sequence is fed back into the RNN, altering its
state and leading to a new prediction for the next step in the sequence.
For molecules it is possible to describe the full configuration of
atoms and connection using the condensed SMILES notation\cite{Weininger1970}.
Letters and special characters are used to describe the topology of
the molecules. Here an RNN is applied to datasets of SMILES strings,
enabling the generation of virtual compound libraries. The molecular
properties of the novel virtual molecules are investigated and compared
with the training datasets and the synthetic feasability evaluated
by scoring and retrosynthetic analysis through computer-aided synthesis
design (CASD).

\section*{Methods}

\subsubsection*{Datasets}

The clean drug like (p13) and the clean fragments subset (p12) was
downloaded in SMILES\cite{Weininger1970} format from the Zinc12\cite{Irwin2005a,Irwin2012}
\href{http://zinc.docking.org/browse/subsets/standard}{website} and
unzipped. The order of the lines was shuffled with the GNU core utilities\cite{Gnu2017}
shuf command line tool. A custom Python\cite{van1995python} script
was used to prefix the SMILES with a start character ``!'' and padded
with an end character ``E'' to a final length two characters longer
than the longest SMILES string in the set. The SMILES sets were subsequently
vectorized by one hot encoding into HDF5\cite{hdf5} data files. The
character-to-index translation information was saved for each dataset.

\subsubsection*{Neural Network}

A RNN was built by using Keras\cite{chollet2015} with Theano\cite{2016arXiv160502688full}
as the computation back end. First two layers were constructed consisting
of 256 LSTM\cite{Hochreiter1997} units run in batch mode with an
input layer matching the number of vectorized characters and the length
of the padded SMILES. This was followed by a time distributed feed
forward network consisting of two hidden layers with 128 neurons with
rectified linear activation. The final output layer matched the number
of characters in the dataset index with a soft-max activation. The
LSTM units were regularized with an input dropout\cite{Srivastava2014}
of 0.1 (dropout\_W).

The neural network was trained by reading chunks of 100.000 vectorized
SMILES into memory which were used in mini-batches of 512 for one
epoch of training for each chunk. The first chunk was reserved for
use as a validation set during training. The learning rate was initially
set to 0.007 but gradually lowered if the validation loss did not
improved within the last five chunks of training.

\subsubsection*{Sampling}

A sampling model was built with the exact same architecture as the
training model, except the LSTM layers were run in stateful mode instead
of batch mode and the input dimension set to a vector size of the
number of characters in the dataset vectorization dictionary. The
output probabilities were adjusted with a sampling temperature, that
rescales the probabilities in the output vector with the following
formula: 
\[
p_{new}=e^{ln(\frac{p_{ori}}{temp})}
\]
Final selection of the next character was then done with NumPy's multinomial
sampler. The networks state was reset before SMILES generation and
the initial character fed to the network was the start character ``!''.
Sampling was terminated when the end character ``E'' was predicted.
The start and end characters were stripped and the SMILES molecular
validity checked by conversion to a sanitized molecule with RDKit\cite{Landrum2016}.

Two datasets of 50.000 molecules were generated with both trained
models at a sampling temperature of 1.0. Using RDKit, they were converted
to canonical SMILES, as were the training sets. The number of similar
molecules was found by intersecting the generated set with the corresponding
full training set.

\subsubsection*{Calculation of Molecular Properties}

The two generated datasets were compared with random samples of 50.000
molecules of both training sets. RDKit was used to calculate a range
of molecular properties: the number of hydrogen bond acceptors, number
of hydrogen bond donors, total polar surface area, number of rotatable
bonds and the molecular weight as well as the predicted LogP\cite{wildman1999prediction}.
In addition, the SA-score\cite{Ertl2009} was calculated on compounds
with the sascorer module from the RDKit SA\_score contribution package.
The compounds were converted to neutral form with the MolVS package\cite{MolVS2017}
before the synthetic accessibility (SA) score was computed. Comparison
plots were made with Matplotlib\cite{Hunter:2007}.

All computations and training were done on a Linux workstation (Ubuntu
Mate 16.04) with 4 GB of RAM, i5-2405S CPU @ 2.50GHz and an Nvidia
Geforce GTX1060 graphics card with 6 GB of RAM.

\subsubsection*{Retrosythetic analysis}

Compounds were selected for retro-synthetic analysis by sampling the
10 compounds nearest the 5, 50 and 95\%th percentile of the SA score
distribution as well as the compound with the highest SA score. Compounds
already found in the training sets were discarded. Each compound was
subjected to retro-synthetic analysis with the Wiley ChemPlanner software\cite{WileyChemplanner2017}
on the default settings. The longest linear route to any intermediate
in the synthesis tree was three steps and common reaction rules were
employed. Reaction rules are defined as common if there are 50 or
more supporting similar examples in the ChemPlanner training database,
uncommon if there are between 25 and 49 supporting similar examples
in the training database, and rare if there are between 5 and 24 supporting
similar examples in the training database. Routes were completed until
commercially available starting materials with a cost less than 1000\$/mol
were found. If it was not possible to generate a route to commercially
available starting materials with the default settings, the maximum
number of linear steps was increased to four, then the sequence three
steps with uncommon rules, three steps with rare rules, four steps
with uncommon rules, and four steps with rare rules was applied until
a solution was found. If no feasible routes were generated, the starting
materials generated in the first round of retrosynthetic planning
were submitted to up to two more retrosynthetic planning queries using
all possible combinations of number of steps and rules.

\section*{Results}

\begin{figure}
\includegraphics[width=1\columnwidth]{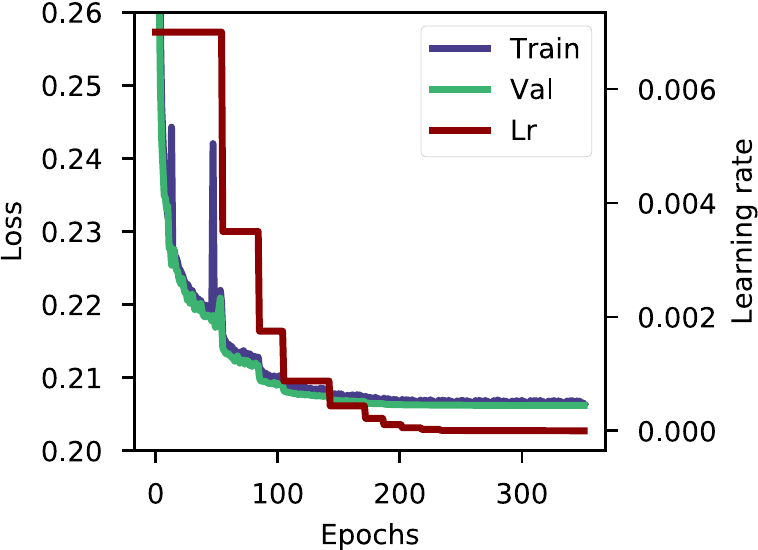}

\caption{\label{fig:Training-History_p12}Training history for the fragment-like
dataset (p12).}

\end{figure}

Training was done for 352 chunks for p12 (fragment-like) and 449 chunks
for p13 (drug-like). The training history for p12 is shown in Figure
\ref{fig:Training-History_p12}.
 The training history for p13 looks essentially similar (not shown),
although a slightly lower final validation loss of 0.167 was obtained.
As p12 contains 1.611.889 SMILES and p13 had 13.195.609, this corresponds
to approximately 22 and 3.5 passes over the entire datasets, respectively.
The difference between the validation loss and training loss were
negligible for both datasets.

\begin{figure}
\includegraphics[width=1\columnwidth]{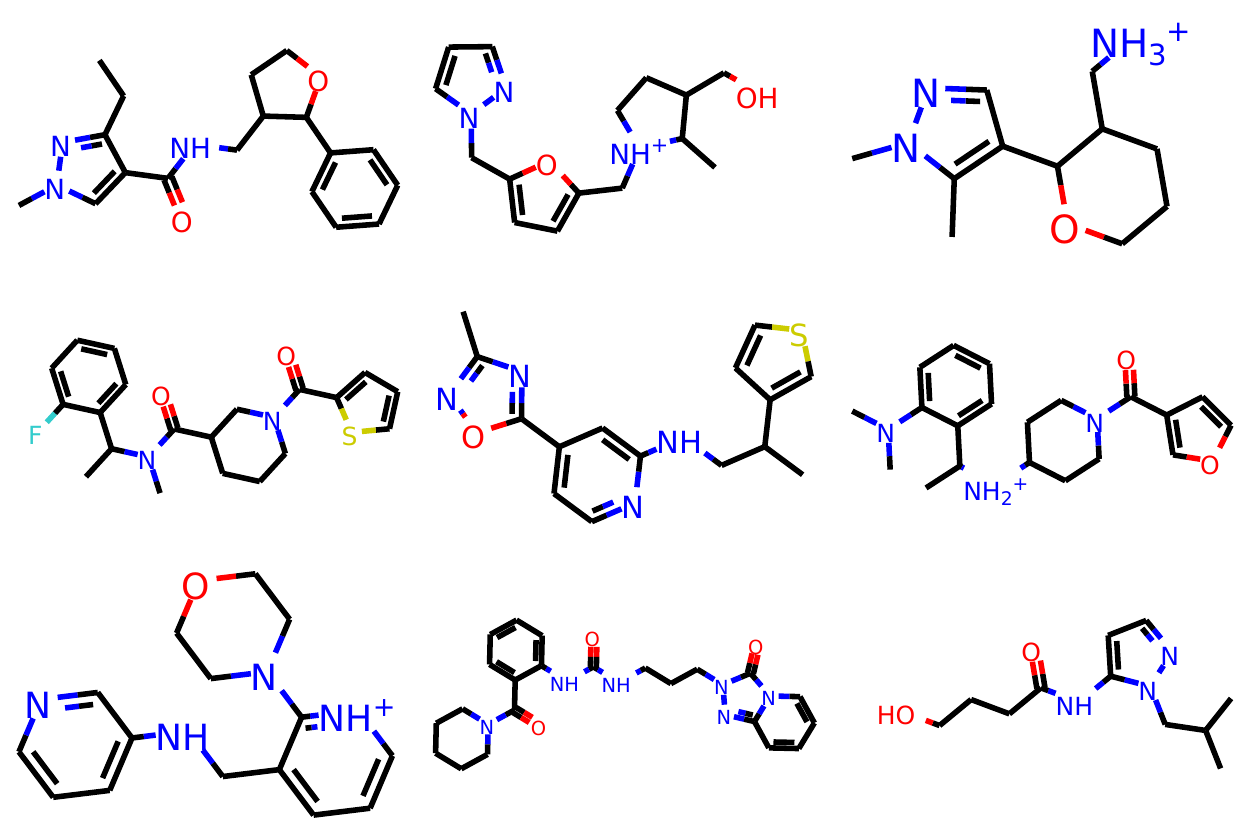}

\caption{\label{fig:Generated_molecules}Examples of generated molecules from
the model trained on drug like molecules (p13).}
\end{figure}

\begin{figure}
\includegraphics[width=1\columnwidth]{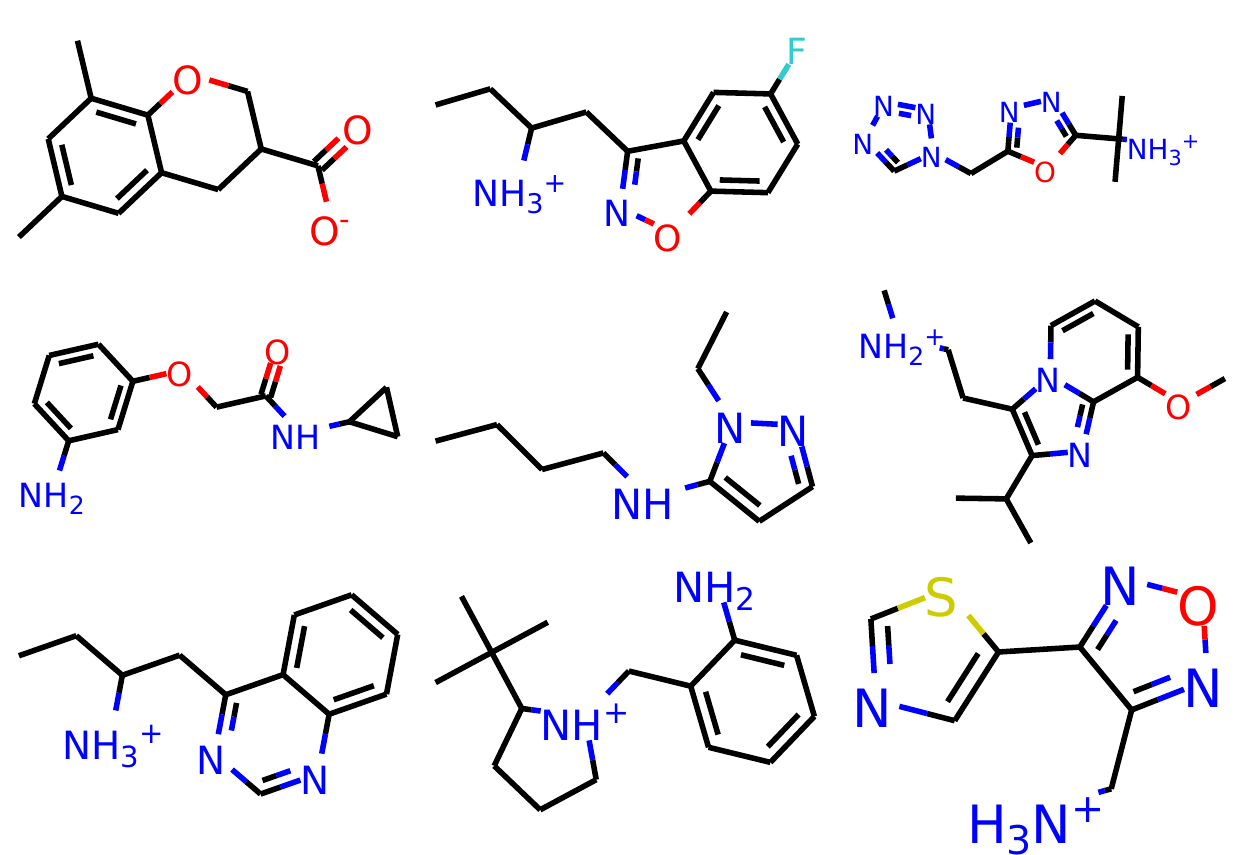}

\caption{\label{fig:Fragment_molecules}Examples of molecules generated from
the model trained on the dataset with molecular fragments (p12).}

\end{figure}

Examples of the molecules generated with the model trained on drug-like
molecules are shown in Figure \ref{fig:Generated_molecules}
. At a glance there seem to be correctly made six membered benzene
rings and some five membered heterocyclic rings which makes electronic
sense according to RDKit. Fused ring systems are also present. Nitrogens
in amides and near aromatic rings appear uncharged, whereas primary
and secondary aliphatic amines are charged, reflecting the charge
normalization of the training set done by the Zinc database. The molecules
generated from the model trained on fragments on average appear to
be smaller and simpler (Figure \ref{fig:Fragment_molecules}
).

\begin{figure}
\begin{centering}
\includegraphics{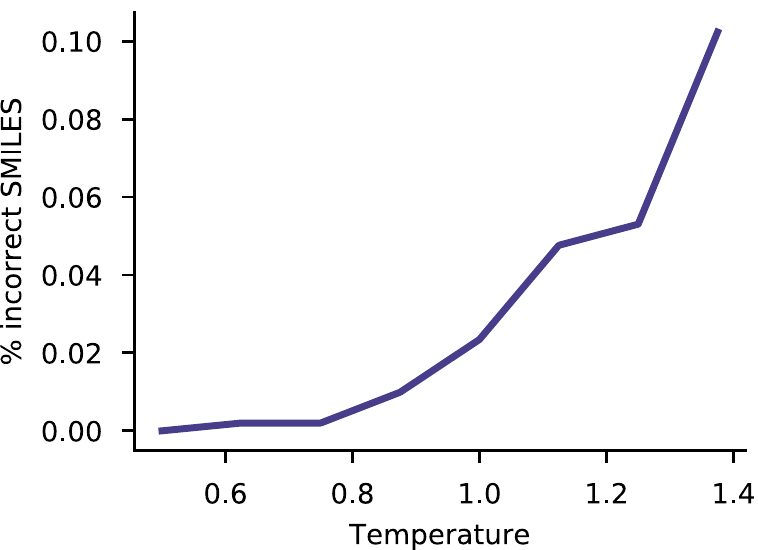}
\par\end{centering}

\caption{\label{fig:Molecular-sampling-error}Molecular sampling error at different
sampling temperature. High sampling temperatures leads to malformed
SMILES strings.}
\end{figure}

Some of the generated SMILES strings were malformed during sampling
and could not be parsed with RDKit. The fraction of the molecules
that contains errors was dependent on the sampling temperature with
approximately 2\% being malformed at a temperature of 1.0 (Figure
\ref{fig:Molecular-sampling-error}
). At temperatures over 1.2 the error rate rises significantly. Typical
errors preventing SMILES parsing were missing closure of parentheses
or unmatched ring closures. Occasionally the valence of atoms was
wrong.

37\% of the generated fragments were found in the training set, whereas
only 17\% of the generated drug-like molecules were found in the training
set.

\subsubsection*{Comparison of molecular features}

\begin{figure*}
\begin{centering}
\includegraphics[width=1\textwidth]{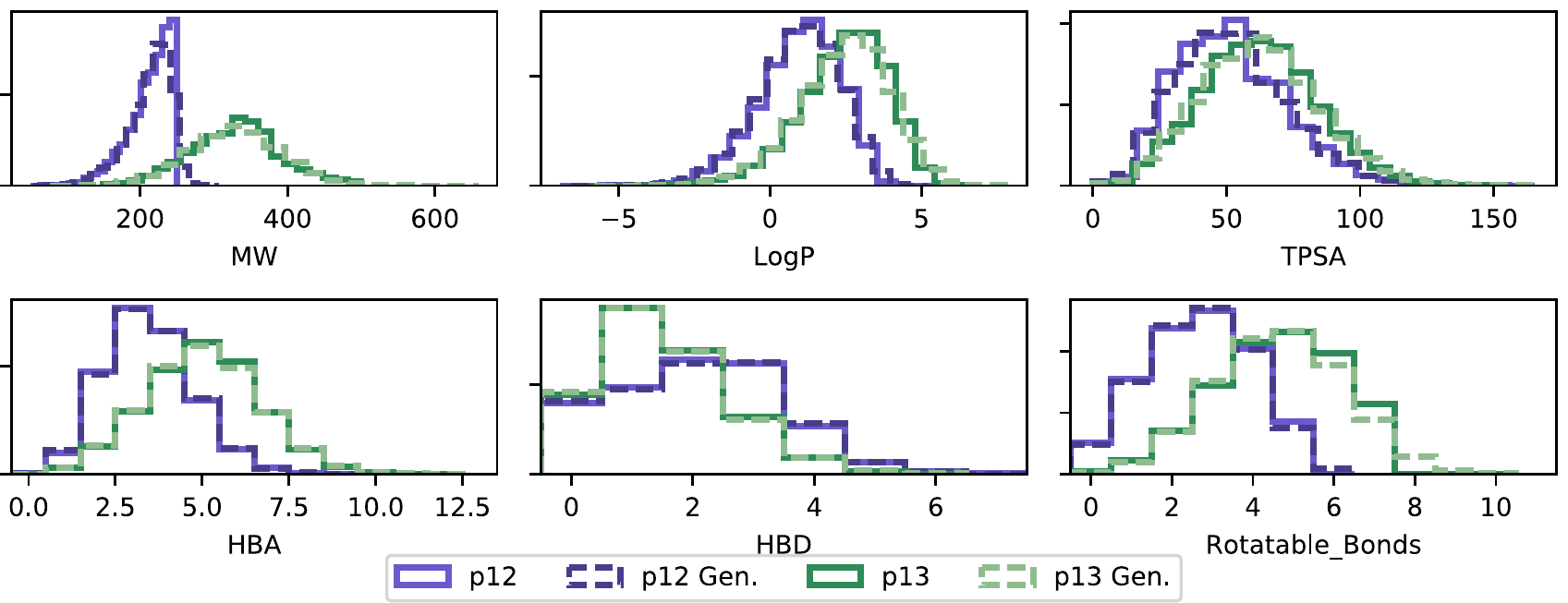}
\par\end{centering}

\caption{\label{fig:Calculated-properties-of}Calculated property distributions
of the generated molecules. The generated molecules distribution of
properties (dashed lines) matches the training sets (solid lines)
used for all calculated properties: Molecular weight (MW), calculated
LogP (LogP), total polar surface area (TPSA), number of hydrogen bond
acceptors (HBA), number of hydrogen bond donors (HBD) and number of
rotatable bonds. p12 and p13 are the Zinc sets for clean fragments
and clean drug like molecules, respectively.}
\end{figure*}

The properties of the molecules generated match the properties of
the molecules used for training of the neural networks. Figure \ref{fig:Calculated-properties-of}
 show the histograms of the properties for both samples of the training
sets and generated sets. The distribution of the properties of the
generated molecules to a large extent overlaps with the distribution
found in the two training sets. The differences between the two training
sets reflects the filtering done by the Zinc database to define the
dataset. The fragments are defined as molecules with xlogp <=3.5 and
mwt <=250 and rotatable bonds <= 5, whereas the drug like molecules
have a molar weight between 150 and 500, xlogp <= 5 and the number
of rotatable bonds <=7. Additionally the drug like molecules must
have a polar surface area below 150, number of hydrogen bond donors
<= 5 and number of hydrogen bond acceptors <= 10. The difference between
the two datasets are most pronounced for the molecular weight histograms,
where the filtering of the training sets are evident as sharp cutoffs
in the histograms. The generated training sets follow this sharp edge
only partially, but they also generate a few molecules with a larger
molecular weight than found in the training sets as evident from the
thin tails.

\begin{figure}
\begin{centering}
\includegraphics[width=1\columnwidth]{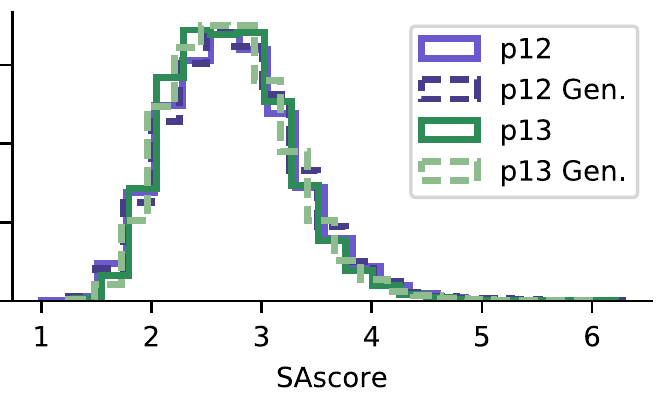}
\par\end{centering}

\caption{\label{fig:SAscore-for-50000}SA score distribution for 50.000 compound
samples of generated molecules and training sets. The SA scores fo
the generated molecules closely follows the SA score of their respective
training sets. p12 and p13 are clean fragments and clean drug-like
molecules respectively.}

\end{figure}

The synthetic feasibility of the generated molecules seem to very
closely follow the training sets as assessed by the SA score\cite{Ertl2009}
(Figure \ref{fig:SAscore-for-50000})
 This score uses a combination of fragment contributions and assessed
topological complexity of the molecule, such as presence of unusual
ring systems, stereo complexity and molecular size\cite{Ertl2009}.
The distributions of the generated and training sets are essentially
similar.

To further investigate the synthesizability of the generated molecules,
three groups of compounds were selected from the drug-like molecules
based on their SA score. A summary of the properties of the identified
synthetic rules for each group are shown in Table \ref{tab:ChemPlanner}.
\begin{table*}
\caption{\label{tab:ChemPlanner}Average properties of the retro-synthetic
routes found by ChemPlanner for three groups sampled near the 5, 50
and 95\%th percentiles of the SA score distribution}

\begin{tabular*}{1\textwidth}{@{\extracolsep{\fill}}c>{\centering}p{2.1cm}c>{\centering}p{2.1cm}cc>{\centering}p{2.5cm}}
\hline 
Group & Possible Selectivity Issues & SA Score & No. of synthesis routes & Max steps & Yield / \% & Cost of SMs US\$/100g\tabularnewline
\hline 
Easy & 0/6 & 1.9 & 26.0 & 2.2 & 55.7 & 1043.8\tabularnewline
Medium & 1/6 & 2.7 & 40.0 & 3.7 & 35.2 & 1690.8\tabularnewline
Hard & 4/9 & 3.7 & 42.1 & 4.4 & 26.1 & 1717.3\tabularnewline
\hline 
\end{tabular*}
\end{table*}
 ChemPlanner generated retrosynthetic routes for all compounds in
the easy group with the default settings. For the medium group, possible
problems were noted for two out of six selected compounds, and the
second highest ranked route was chosen for one of them. A route to
one of the compounds in the hard group could not be established, even
though the number of steps was and more rare reaction rules were applied.
Moreover, for the compound with the highest SA score, no route was
identified from commercially available starting materials. The summary
in Table \ref{tab:ChemPlanner} illustrates the expected differences
between the groups. The highest proportion of possible selectivity
issues was found in the group with the highest SA score and this group
required the most steps, would be expected to give the lowest overall
yield and had the highest average cost of the starting materials.
The number of synthetic routes was larger for the hard group possibly
reflecting the larger amount of combinations for longer synthetic
routes with more intermediates. The full table with details on each
sampled compound can be found in the Supplementary Information.

\section*{Discussion}

The most obvious question to ask of the automatically created novel
molecules is regarding their practical synthesizability. The created
compounds complexity, topology and created fragments closely matches
the ones found in the training set, which are already synthesized
compounds. The calculated SA scores thus seem to match very closely
and is in the easy to medium end of the range and matches the SA scores
observed for catalog compounds in the original study\cite{Ertl2009}.
Using Wiley ChemPlanner\cite{WileyChemplanner2017}, three groups
of molecules were subjected to retrosynthetic planning. For the majority
of compound retrosynthetic routes could be identified (c.f. Table
\ref{tab:ChemPlanner}). There seemed to be a larger proportion of
possible problems with compounds from the group with the highest SA
scores, indicating that the SA score could be used as a first screen
of the generated compound before further detailed evaluation if synthesizability
and cost is an issue. The SA scores are however in the medium to low
range when compared to the scores reported\cite{Ertl2009}, illustrating
that for all except the easiest cases, the compounds should be more
detailed evaluated by retro-synthetic software\cite{WileyChemplanner2017}
operated by trained medicinal chemists.

The sampling provided novel molecules, 63\% and 83\% which were not
found in the training set, for the fragment-like and drug-like sets,
respectively. The larger drug-like molecules were more novel, even
if the training set was almost an order of magnitude larger and thus
represents a larger sampling set in which to re-find a given molecule.
This is possibly due to a larger possible combinatorial space for
the atomic connections of the molecules. Considering the large estimated
numbers of drug like molecules (see above), it is surprising that
the recurrent network generates such a large proportion of molecules
already found in the training sets. This indicates over fitting of
the networks, which is not apparent from the training graphs, Figure
\ref{fig:Training-History_p12}, and the similarity of the final loss
function from the training and test set. This could reflect that the
training and test sets are too similar in properties: The original
database could contain a high proportion of compound series that have
been split into both the training and test sets. This effect could
possibly be mitigated by performing a clustering based on scaffolds
as encoded by circular fingerprints followed by a division into training
and test set based on clusters. This could ensure a setting of hyperparameters
that makes the networks cover more diverse areas of chemical space.
As it is now, the networks seem to only generates molecules in the
vicinity of the training molecules, leading only to a small expansion
of the chemical space covered. 

On the other hand, this propensity to generate molecules in the vicinity
of the training sets could be used to generate molecules similar to
compound with known biological activity. However, the small size of
the datasets with known actives easily leads to extensive overfitting,
although limited retraining of previously trained networks seems promising\cite{MolCreativ2016}.
Techniques such as data augmentation with SMILES enumeration could
also be of potential use\cite{Bjerrum2017}. A recent publication
used a similar approach with reinforcement learning to tune a generative
RNN for producing molecules and additionally demonstrated how the
network could be tuned towards generating molecules that potentially
have bioactivity\cite{Olivecrona2017}.

The close link between the training set and the generated molecules
could potentially lead to interesting iterative approaches in \textit{in
silico} drug discovery resembling genetic algorithms. The network
could be used to generate a set of molecules which are subsequently
evaluated with QSAR models or molecular docking. The best half is
then used to retrain the network. After a couple of cycles the produced
molecules would likely have better properties in the \textit{in silico}
models.

\section*{Conclusion}

Recurrent neural networks with LSTM cells can be trained to generate
novel and chemically plausible molecules as assessed by SA score and
retro-synthetic analysis with Wiley ChemPlanner. The distribution
of the molecular properties of the generated molecules closely resembles
the property distributions in the used training sets, making it possible
to tune the networks by filtering the datasets. This possibility leads
to novel opportunities for expanding known compound series with similar
compounds of matched properties for generation of larger \textit{in
silico} compound libraries. There was a correlation between the SA
score and the properties of the retrosynthetic routes, although no
routes could be generated for two of the compounds selected for detailed
analysis.

\section*{Conflict of interests}

E. J. Bjerrum is the owner of Wildcard Pharmaceutical Consulting.
The company is usually contracted by biotechnology/pharmaceutical
companies to provide third party services. R. Threlfall is an employee
of Wiley-VCH.

\bibliographystyle{elsarticle-num}
\bibliography{MolGeneration_db.bib}

\end{document}